\documentclass[preprint, 12pt]{elsarticle}

\usepackage{lineno, hyperref}
\usepackage{booktabs}
\usepackage{amssymb}
\usepackage{makecell}
\usepackage{subfigure}
\usepackage{caption}
\usepackage{amsmath}
\usepackage{color, soul}
\usepackage{threeparttable}
\usepackage{arydshln}

\journal{Journal of Systems Architecture}

\begin{document}
\begin{frontmatter}



\title{Post-Training Quantization for Re-parameterization via Coarse \& Fine Weight Splitting}


\author[org]{Dawei Yang\corref{cor1}}
\author[org,org2]{Ning He\corref{cor1}}

\author[org]{Xing Hu}

\author[org]{Zhihang Yuan\corref{cor2}}
\ead{hahnyuan@gmail.com}

\author[org]{Jiangyong Yu}

\author[org]{Chen Xu}

\author[org1,org2]{Zhe Jiang\corref{cor2}}
\ead{zj266@cam.ac.uk}

\cortext[cor1]{These authors contributed equally to this work}
\cortext[cor2]{Corresponding author}

\affiliation[org]{organization={Houmo AI},
            country={China}}

\affiliation[org1]{organization={University of Cambridge},
            country={United Kingdom}}

\affiliation[org2]{organization={Southeast University},
            country={China}}

\begin{abstract}

Although neural networks have made remarkable advancements in various applications, they require substantial computational and memory resources. Network quantization is a powerful technique to compress neural networks, allowing for more efficient and scalable AI deployments.
Recently, Re-parameterization has emerged as a promising technique to enhance model performance while simultaneously alleviating the computational burden in various computer vision tasks.
However, the accuracy drops significantly when applying quantization on the re-parameterized networks. We identify that the primary challenge arises from the large variation in weight distribution across the original branches.
To address this issue, we propose a coarse \& fine weight splitting (CFWS) method to reduce quantization error of weight, and develop an improved KL metric to determine optimal quantization scales for activation.
To the best of our knowledge, our approach is the first work that enables post-training quantization applicable on re-parameterized networks. For example, the quantized RepVGG-A1 model exhibits a mere 0.3\% accuracy loss. The code is in https://github.com/NeonHo/Coarse-Fine-Weight-Split.git
\end{abstract}






\end{frontmatter}


\section{Introduction}
While CNNs have demonstrated dominant performance across a wide range of applications, it is still expected to perform efficiently on both edge devices and cloud servers.
Neural network quantization \cite{gholami2021survey, nagel2021white, wu2020integer, krishnamoorthi2018quantizing, esser2019learned, hubara2021accurate} is one of the most widely used techniques to compress the neural network. It involves reducing the precision of network parameters and activations, typically from floating-point numbers (e.g., FP32) to integers with lower bit (e.g., INT8). It greatly reduces memory cost (e.g., INT8 saves 75\% model size and bandwidth) and accelerates matrix multiplication (e.g., Conv, FC) due to integer computation. Network quantization is typically classified into two categories: post-training quantization (PTQ) \cite{nagel2021white, hubara2021accurate} and quantization-aware training (QAT) \cite{krishnamoorthi2018quantizing, esser2019learned}. PTQ takes a trained network and quantize it with little or no data, thus it requires minimal hyper-parameters tuning and no end-to-end training. In contrast, QAT is done in the model training or re-training process with simulated quantization. So it usually requires substantial computational resources and more effort in training (e.g., hyper-parameter tuning), leading to a complex deployment procedure. In addition, when deploying neural networks, accessing to training resources (e.g., training data, training code) may be restricted or limited due to privacy concern or data security \cite{nagel2021white}. Thus, in this paper, our focus is specifically on applying PTQ to re-parameterized networks.



Recently, re-parameterization \cite{ding2021repvgg} has emerged as a promising technique and has also been widely used in various tasks \cite{ding2022repmlpnet,ding2021resrep,ding2022scaling,li2022yolov6,wang2022yolov7}. It involves substituting each linear layer, such as FC and Conv, with a block of linear layers. This kind of block is only used during training, and will be ultimately merged into a single linear layer during inference.
On one hand, re-parameterization leverages the exceptional performance of their multi-branch architecture during training. On the other hand, the converted plain single-branch structure capitalizes on their high parallelism and reduced memory footprint, resulting in improved inference efficiency. For example, the pioneer work RepVGG \cite{ding2021repvgg} reaches over 80\% top-1 accuracy on ImageNet and runs 83\% faster than ResNet-50 \cite{he2016deep} with higher accuracy. It is known that the residual path of ResNet requires on-the-fly memory retention, which can be challenging for resource-constrained edge devices.
However, the accuracy drops significantly when directly applying quantization methods on re-parameterized networks. For example, the accuracy of RepVGG-A1 drops from $74.5\%$ to $61.7\%$ after PTQ. The accuracy degradation of quantized reparameterization models poses a hindrance to their practical application. It is worth mentioning that due to the absence of the BN layer in the deploy mode of re-parameterized networks, it is also difficult to enhance quantization accuracy through QAT.

As of our current knowledge, there are only two works \cite{ding2022re, chu2022make} that explore quantization of re-parameterized networks. However, despite their novelty, both approaches require modifications to the original re-parametrization structure. Specifically, \cite{chu2022make} moves BN layer outside the rep-block; \cite{ding2022re} drops the re-parameterization structure but instead of adding re-parameterization into the optimizer. While the converted single-branch maintains the same inference-time structure, the training-time process becomes significantly more intricate, requiring additional time, steps and adjustments in the training pipeline. 
Besides, both approaches require training the modified model from scratch, which is a significant limitation.
\textbf{In contrast, our objective is to ensure that model research and model deployment remain independent:} (1) allowing researchers to focus on improving the structure and accuracy of floating-point models without being hindered by the quantization process. (2) simplifying the deployment process. The second point is crucial because training models from scratch may not always be practical, especially in scenarios where there is a lack of dataset availability (e.g., when the model is provided by a vendor or when the dataset is confidential), a need for rapid deployment, or constraints in terms of training costs.

\begin{figure*}
    \centering
    \setlength{\tabcolsep}{0pt}
    \resizebox{1.0\textwidth}{!}{
    \begin{tabular}{cccc:c}
        \scriptsize{9th} &  
        \begin{minipage}{0.2\textwidth}
        \centering
        \includegraphics[scale=0.8,  trim=8 5 0 19, clip]{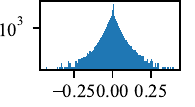}
        \end{minipage} &
        \begin{minipage}{0.2\textwidth}
        \centering
        \includegraphics[scale=0.8,  trim=8 5 0 19, clip]{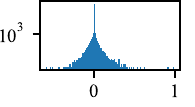}
        \end{minipage} &
        \begin{minipage}{0.2\textwidth}
        \centering
        \includegraphics[scale=0.8,  trim=8 5 0 19, clip]{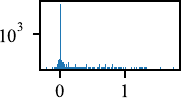}
        \end{minipage}&
        \begin{minipage}{0.2\textwidth}
        \centering
        \includegraphics[scale=0.8,  trim=8 5 0 19, clip]{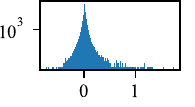}
        \end{minipage}\\
        \scriptsize{10th}&  
        \begin{minipage}{0.2\textwidth}
        \centering
        \includegraphics[scale=0.8,  trim=8 5 0 19, clip]{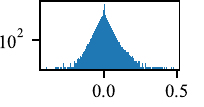}
        \end{minipage}&
        \begin{minipage}{0.2\textwidth}
        \centering
        \includegraphics[scale=0.8,  trim=8 5 0 19, clip]{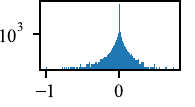}     
        \end{minipage}&
        \begin{minipage}{0.2\textwidth}
        \centering
        \includegraphics[scale=0.8,  trim=8 5 0 19, clip]{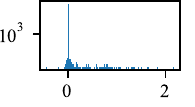}
        \end{minipage}&
        \begin{minipage}{0.2\textwidth}
        \centering
        \includegraphics[scale=0.8,  trim=8 5 0 19, clip]{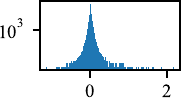}
        \end{minipage}\\
        \scriptsize{11st}&  
        \begin{minipage}{0.2\textwidth}
        \centering
        \includegraphics[scale=0.8,  trim=8 5 0 19, clip]{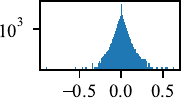}
        \end{minipage}&
        \begin{minipage}{0.2\textwidth}
        \centering
        \includegraphics[scale=0.8,  trim=8 5 0 19, clip]{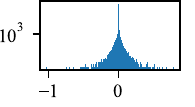}     
        \end{minipage}&
        \begin{minipage}{0.2\textwidth}
        \centering
        \includegraphics[scale=0.8,  trim=8 5 0 19, clip]{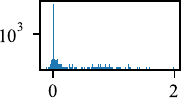}
        \end{minipage}&
        \begin{minipage}{0.2\textwidth}
        \centering
        \includegraphics[scale=0.8,  trim=8 5 0 19, clip]{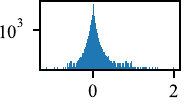}
        \end{minipage}\\
        \scriptsize{12nd}&  
        \begin{minipage}{0.2\textwidth}
        \centering
        \includegraphics[scale=0.8,  trim=8 5 0 19, clip]{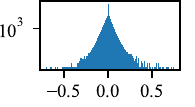}
        \end{minipage}&
        \begin{minipage}{0.2\textwidth}
        \centering
        \includegraphics[scale=0.8,  trim=8 5 0 19, clip]{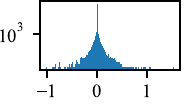}     
        \end{minipage}&
        \begin{minipage}{0.2\textwidth}
        \centering
        \includegraphics[scale=0.8,  trim=8 5 0 19, clip]{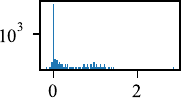}
        \end{minipage}&
        
        \begin{minipage}{0.2\textwidth}
        \centering
        \includegraphics[scale=0.8,  trim=8 5 0 19, clip]{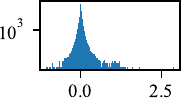}
        \end{minipage}\\
        \scriptsize{13rd}& 
        \begin{minipage}{0.2\textwidth}
        \centering
        \includegraphics[scale=0.8,  trim=8 5 0 19, clip]{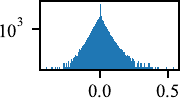}
        \end{minipage}&
        \begin{minipage}{0.2\textwidth}
        \centering
        \includegraphics[scale=0.8,  trim=8 5 0 19, clip]{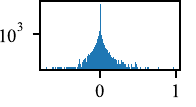}     
        \end{minipage}&
        \begin{minipage}{0.2\textwidth}
        \centering
        \includegraphics[scale=0.8,  trim=8 5 0 19, clip]{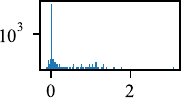}
        \end{minipage}&
        
        \begin{minipage}{0.2\textwidth}
        \centering
        \includegraphics[scale=0.8,  trim=8 5 0 19, clip]{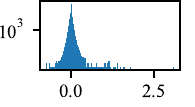}
        \end{minipage}\\
        \scriptsize{14th}& 
        \begin{minipage}{0.2\textwidth}
        \centering
        \includegraphics[scale=0.8,  trim=8 5 0 19, clip]{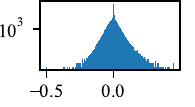}
        \end{minipage}&
        \begin{minipage}{0.2\textwidth}
        \centering
        \includegraphics[scale=0.8,  trim=8 5 0 19, clip]{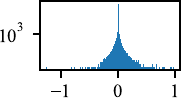}     
        \end{minipage}&
        \begin{minipage}{0.2\textwidth}
        \centering
        \includegraphics[scale=0.8,  trim=8 5 0 19, clip]{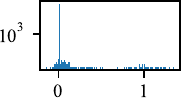}
        \end{minipage}&
        
        \begin{minipage}{0.2\textwidth}
        \centering
        \includegraphics[scale=0.8,  trim=8 5 0 19, clip]{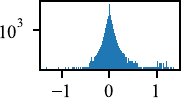}
        \end{minipage}\\
         & \scriptsize{3x3 Conv Branch} 
         & \scriptsize{1x1 Conv Branch} 
         & \scriptsize{Identity Branch} 
         & \scriptsize{Converted Branch}\\
    \end{tabular}
    }
    \caption{Weight distribution of $9^{th}\sim 14^{th}$ re-parameterization blocks before conversion (3 columns on the left) and after conversion (last column on the right). The weight distribution across different branches exhibits substantial disparities, leading to a broader range of converted weights. Besides, a significant portion of values cluster around 0, while values that deviate from 0 also hold considerable importance (i.e., mainly derived from 1x1 and identity branches). This combination of concentrated values around 0 and the significance of values far from 0 presents a challenging scenario for quantification.}
    \label{tab:dists}
\end{figure*}

In order to apply PTQ on re-parameterized networks without significant accuracy degradation, we investigated this task to identify the primary challenge. We have discovered that the quantization difficulty is caused by the large variation in weight distribution across the original branches, resulting in an unfavorable parameter distribution of the fused kernels. As shown in Figure \ref{tab:dists}, the weight distribution across different branches exhibits substantial disparities, leading to a broader range of converted weights. For instance, in the 12nd conv branch, the ranges of 3x3 conv, 1x1 conv, and identity branch are approximately [-0.8, 0.8], [-1.2, 1.8], and [-0.2, 2.8] respectively; the final range extends to [-1.2, 2.8] after conversion. Upon observation, it becomes apparent that a significant portion of values cluster around 0, while values that deviate from 0 also hold considerable importance (i.e., mainly derived from 1x1 and identity branches). This combination of concentrated values around 0 and the significance of values far from 0 presents a challenging scenario for quantification.

We further investigate the problem based on the conversion process (as illustrated in Figure \ref{fig:conversion_kernels}) where the weights from the identity branch and the 1×1 Conv only influence the central points within the merged 3×3 kernel. Specifically, the center points of the converted kernel exhibit a wide range of values, while the surrounding weights (i.e., the eight non-central points in a 3×3 matrix) have a smaller range (for more details, refer to Figure \ref{fig:boxplot}). \cite{ding2022re} also indicates that the central points of the kernel have a higher standard deviation compared to the surrounding points. Therefore, our focus should be on properly quantizing the center points within the converted kernel. We propose Coarse \& Fine Weight Splitting (CFWS) to quantize the center weights and surrounding weights separately.
In terms of activation quantization, we have notices that many large activation values are insensitive to clipping. This observation motivates us to propose a calibration metric based on KL divergence \cite{joyce2011kullback}. The improved metric enables appropriate clipping of the large activation values, thereby mitigating quantization errors effectively.
We evaluate our proposed PTQ method on various re-parameterized networks designed for different tasks. The results show that our method is not only effective for image classification but also exhibits promising generalization capabilities for object detection tasks.
Our contributions are as follows:
\begin{enumerate}
\item[1)] We propose a novel PTQ method for re-parameterized networks, without requiring any modifications to the network or re-training. Our method effectively bridges the gap between re-parametrization techniques and the practical deployment of PTQ.
\item[2)] We analysis the underlying cause of performance degradation in the quantization of reparameterization-based architectures, and propose a novel technique called Coarse \& Fine Weight Splitting (CFWS). This approach enables the separate quantization of center weights and surrounding weights, effectively addressing the issue.
\item[3)] We develop an improved KL metric that smoothly handles the proper clipping of large activation values. Our overall method generalizes well on various vision tasks and achieves competitive quantization accuracy, meanwhile retaining the advantages of rapid PTQ deployment.
\end{enumerate}

\begin{figure}
    \centering
    \includegraphics[width=1.0\linewidth]{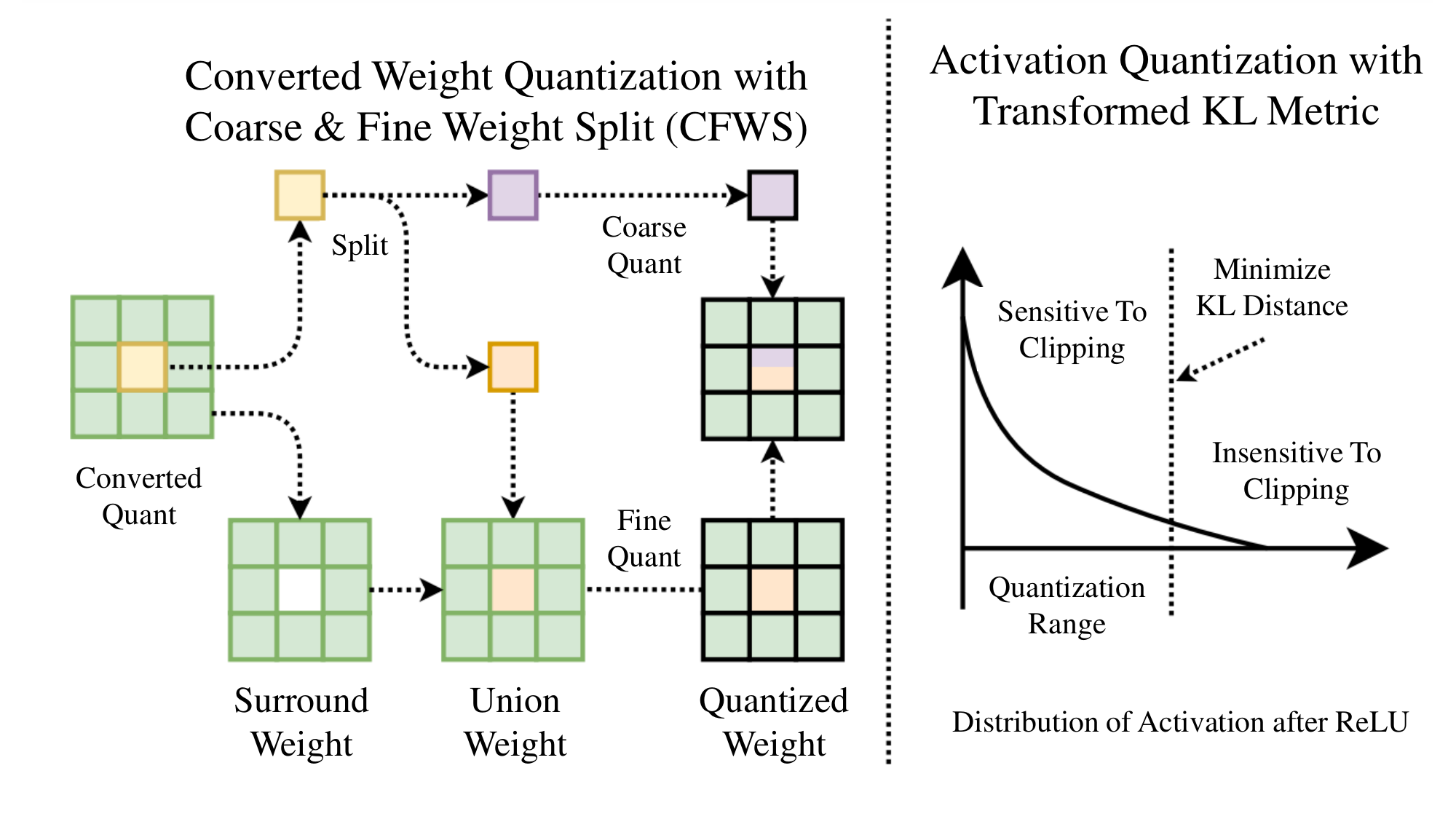}
    \caption{Overview of our proposed PTQ method on re-parameterized networks. The left part is the coarse \& fine weight quantization, and the right part is the optimized KL metric for PTQ calibration.}
    \label{fig:my_label}
\end{figure}

\section{Background and Related Work}


\subsection{Quantization}\label{sec:quant_quant_friendly}

Neural network quantization is a technique utilized in deep learning to reduce the memory and computational requirements of neural networks \cite{gholami2021survey, nagel2021white, wu2020integer, krishnamoorthi2018quantizing, esser2019learned, hubara2021accurate}. It involves representing the weights and activations of a neural network using fewer bits compared to the standard floating-point representation. Typically, deep neural networks store weights and activations as 32-bit floating-point numbers, but using lower-precision representations can significantly reduce the memory usage and improve computational efficiency. For instance, replacing 32-bit floating-point (FP32) with 8-bit integers (INT8) can result in a 75\% reduction in model size and bandwidth, as well as accelerate matrix multiplication (e.g., Conv, FC) due to integer computation.

Network quantization can be categorized into two main approaches: post-training quantization (PTQ) \cite{nagel2021white, hubara2021accurate} and quantization-aware training (QAT) \cite{krishnamoorthi2018quantizing, esser2019learned}. PTQ involves quantizing a pre-trained network with minimal or no data, requiring minimal hyper-parameter tuning and no end-to-end training. On the other hand, QAT is performed during the model training or re-training process by simulating quantization. QAT typically demands substantial computational resources and more training effort, such as hyper-parameter tuning, resulting in a complex deployment procedure. Additionally, deploying neural networks may encounter restrictions or limitations on accessing the training data due to privacy or data security concerns \cite{nagel2021white}. Since QAT involves modifying the training code and entails additional costs, it is only employed when the training resources (code and data) are available and PTQ fails to produce satisfactory results. In this paper, our specific focus is on applying PTQ to re-parameterized networks.

\subsection{Re-parameterization}

Re-parametrization are not a novel concept and can be traced back to batch normalization \cite{ioffe2015batch} and residual connection \cite{he2016deep}. A recent contribution by Ding et al. \cite{ding2021repvgg} introduced the RepVGG network architecture, which utilizes structure re-parameterization. Since then, re-parametrization has shown promise and has been widely adopted in various tasks \cite{ding2022repmlpnet, ding2021resrep, ding2022scaling, li2022yolov6, wang2022yolov7, wang2022repsr}. Basically, it involves substituting each linear layer, such as FC and Conv, with a block of linear layers. This kind of block is only used during training, and will be ultimately merged into a single linear layer during inference. The converted network achieves the same prediction accuracy as the network during training.

Re-parameterization offers several advantages. Firstly, it leverages the superior performance and faster convergence of the multi-branch architecture. Secondly, the converted plain single-branch structure takes advantage of high parallelism and reduced memory footprint, resulting in improved inference efficiency. For instance, the pioneering work RepVGG \cite{ding2021repvgg} achieves over 80\% top-1 accuracy on ImageNet and runs 83\% faster than ResNet-50 \cite{he2016deep} or 101\% faster than ResNet-101 with higher accuracy. It is known that the residual path of ResNet requires on-the-fly memory retention, which can be challenging for resource-constrained edge devices. Although re-parameterization offers clear advantages in terms of both accuracy and speed, the accuracy degradation of quantized re-parameterization models poses a hindrance to their practical application \cite{chu2022make}.

\subsection{Exploration on Quantizing Re-parametrization Networks}

Reparameterization-based architectures pose challenges in quantization due to their inherent multi-branch design, which leads to an increased dynamic numerical range. Currently, only two works \cite{ding2022re, chu2022make} have explored the quantization of re-parameterized networks. Ding et al. \cite{ding2022re} introduced the gradient re-parameterization (GR) scheme, which replaces re-parameterization by incorporating prior knowledge into a network-specific optimizer called RepOptimizer. Chu et al. \cite{chu2022make} redesigned RepVGG as QARepVGG, which generates weight and activation distributions optimized for quantization. However, both approaches require modifications to the original re-parameterization structure. Specifically, \cite{chu2022make} moves the BN layer outside the rep-block, while \cite{ding2022re} eliminates the re-parameterization structure and integrates it into the optimizer. Although the converted single-branch maintains the same inference-time structure, the training process becomes significantly more complex, necessitating additional time, steps, and adjustments in the training pipeline. In contrast, our method maintains the independence of model research and PTQ deployment.

\section{Method}

The quantization function is to transform a floating-point vector $X$ into a fix-point value $X^q$.
The symmetric uniform quantization $Q$ is the most common method \cite{gholami2021survey}, which is formulated as:
\begin{equation}
X^q = Q(X,s) = \text{clip}(\text{round}(\frac{X}{s}),-2^{k-1},2^{k-1}-1)
\end{equation}
where $k$ is the bit-width of quantization (in general, k=8), and $s$ is a quantization parameter named scale factor. The dequantization of $X^q$ is represented as $\mathcal{X} = X^q \cdot s$.
Quantization errors $X - \mathcal{X}$ typically have two sources: the error resulting from the rounding function is known as rounding error, while the error arising from the clip function is referred to as clipping error. Values outside the quantization range are referred to as outliers and will be clipped.

When appling quantization on a neural network, the closeness of performance to the floating-point network is the main evaluation criterion for assessing the effectiveness of the quantization method. A general quantization objective is:
\begin{equation}
\underset{s}{\operatorname{argmin}}\ L(X, \mathcal{X})
\end{equation}
where $L$ is an objective function such as Mean Squared Error (MSE). Our goal is to find appropriate values for $s$ and $L$ to maximize the performance of quantized network without incurring additional retraining costs.

\begin{figure}
    \centering
    \includegraphics[trim=0 0 0 0, clip]{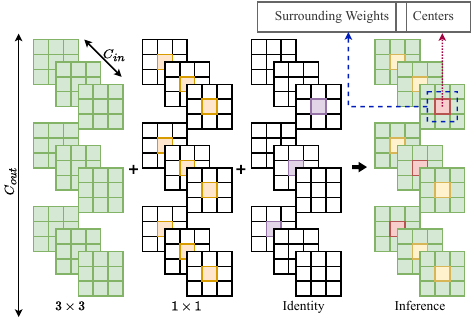}
    \caption{The conversion process of a re-parameterization block. The identity branch and the 1×1 conv only influence the central points within the merged 3×3 kernel.}
    \label{fig:conversion_kernels}
\end{figure}

\begin{figure}
    \centering
    \subfigure[$19^{th}$]{
        \begin{minipage}{0.3\textwidth}
            \centering
            \includegraphics[scale=0.55, trim=12 0 0 19, clip]{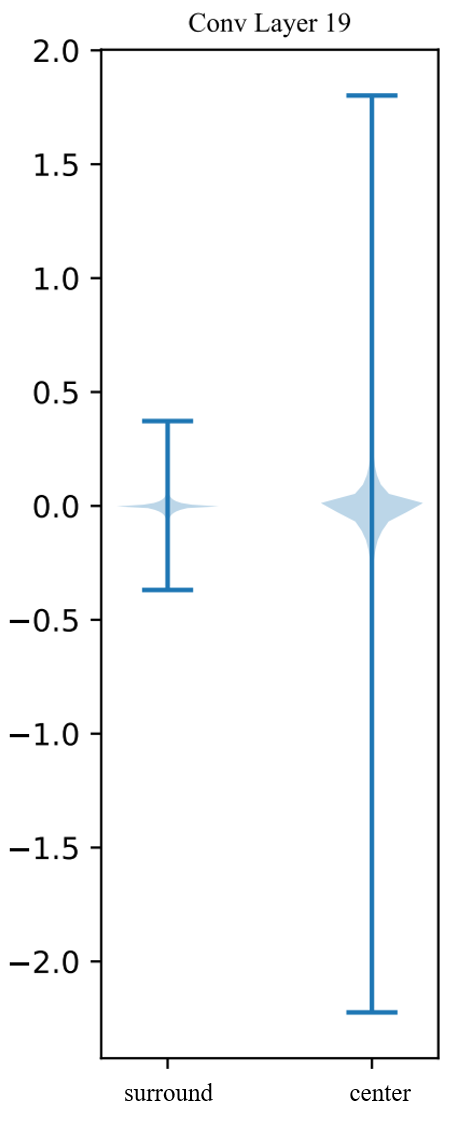}
        \end{minipage}\label{fig:boxplota}
    }
    \subfigure[$20^{th}$]{
        \begin{minipage}{0.3\textwidth}
            \centering
            \includegraphics[scale=0.55, trim=12 0 0 22, clip]{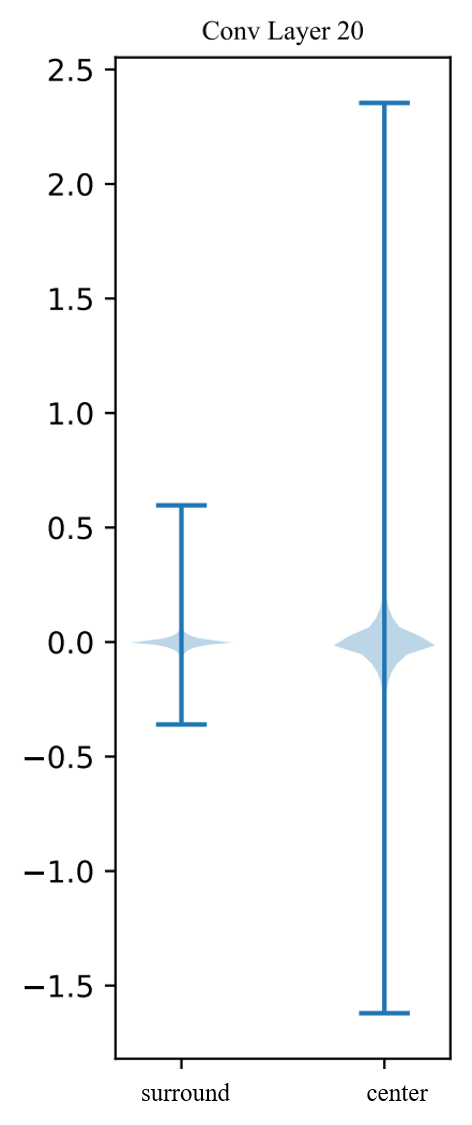}
        \end{minipage}\label{fig:boxplotb}
    }
    \subfigure[$21^{st}$]{
        \begin{minipage}{0.3\textwidth}
            \centering
            \includegraphics[scale=0.55, trim=12 0 0 22, clip]{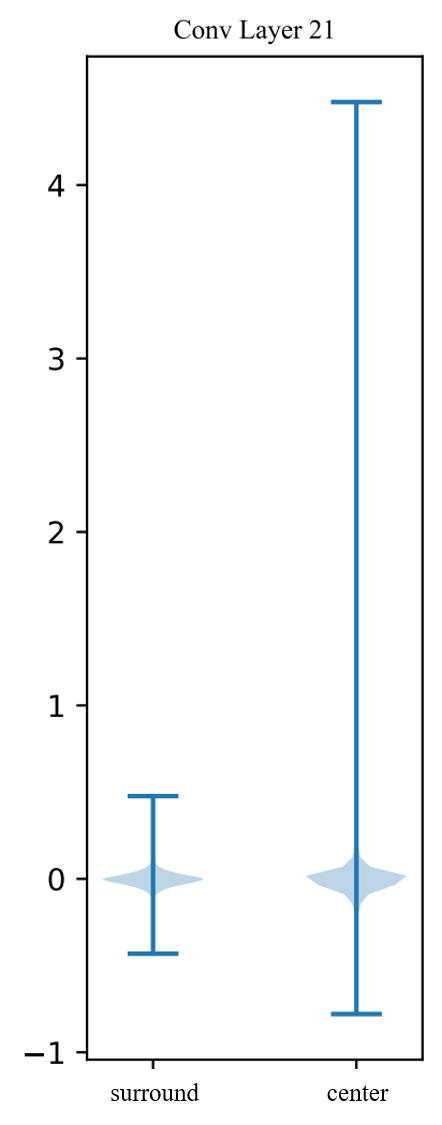}
        \end{minipage}\label{fig:boxplotc}
    }
    \caption{The distribution of the surrounding (non-central) weights and the center weights of $19^{th}\sim 21^{st}$ layers. The y-axis represents the magnitude of the weights.}
    \label{fig:boxplot}
\end{figure}

In the following subsections, we first analyze the underlying causes of performance degradation in the quantization of reparameterization-based architectures, and then introduce our PTQ method for both weight and activation quantization in reparameterization networks.

\subsection{Coarse \& Fine Weight Splitting}

To facilitate understanding, we have visualized the weight distribution of the RepVGG block before and after conversion in Figure \ref{tab:dists}. The weight distribution across different branches exhibits substantial disparities, leading to a broader range of converted weights. For instance, in the 12nd conv branch, the ranges of 3x3 conv, 1x1 conv, and identity branch are approximately [-0.8, 0.8], [-1.2, 1.8], and [-0.2, 2.8] respectively; the final range extends to [-1.2, 2.8] after conversion. Upon observation, it becomes apparent that a significant portion of values cluster around 0, while values that deviate from 0 also hold considerable importance (i.e., mainly derived from 1x1 and identity branches). This combination of concentrated values surrounding 0 and the significance of values far from 0 presents a challenging scenario for quantification. \cite{chmiel2020robust} indicates that weights following a uniform distribution are more resilient to quantization. In our case, the direct application of PTQ methods like MinMax can result in an overly wide quantization range. This wide range can lead to the loss of a significant portion of weights concentrated in a small segment during quantization, thereby hindering the conv layers' ability to preserve their original feature extraction capabilities.

\begin{figure}
    \centering
    \subfigure[$19^{th}$]{
        \begin{minipage}{0.18\textwidth}
            \centering
            \includegraphics[scale=0.6, trim=20 0 0 17, clip]{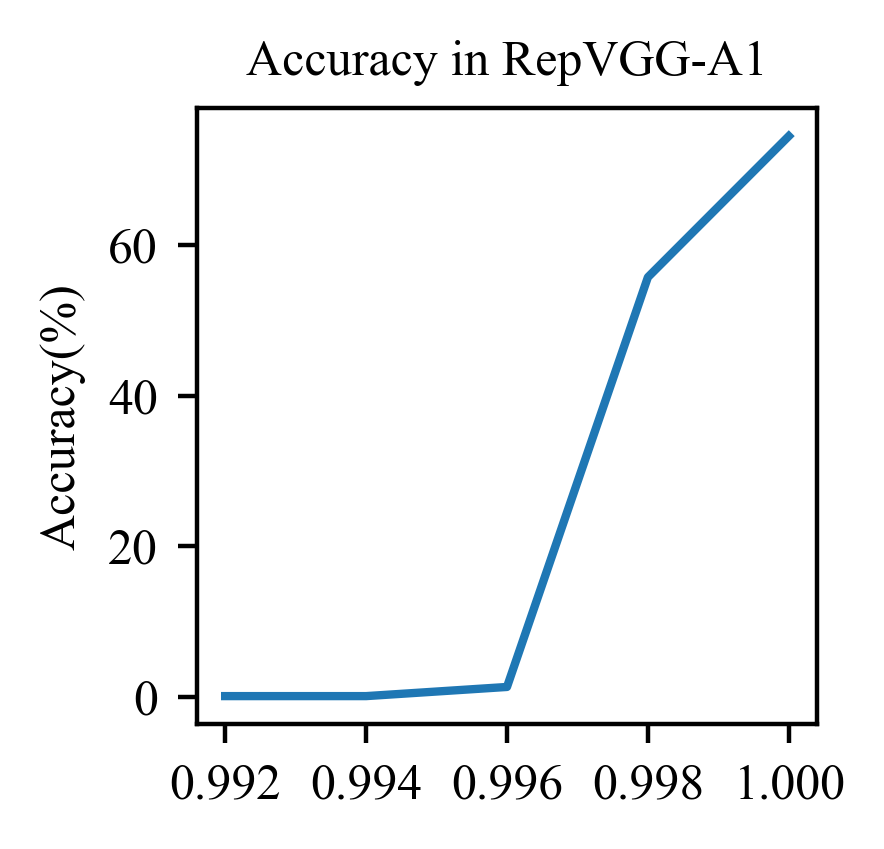}
        \end{minipage}\label{fig:repvgg_a1_w_lines21}
    }
    \subfigure[$20^{th}$]{
        \begin{minipage}{0.18\textwidth}
            \centering
            \includegraphics[scale=0.6, trim=20 0 0 17, clip]{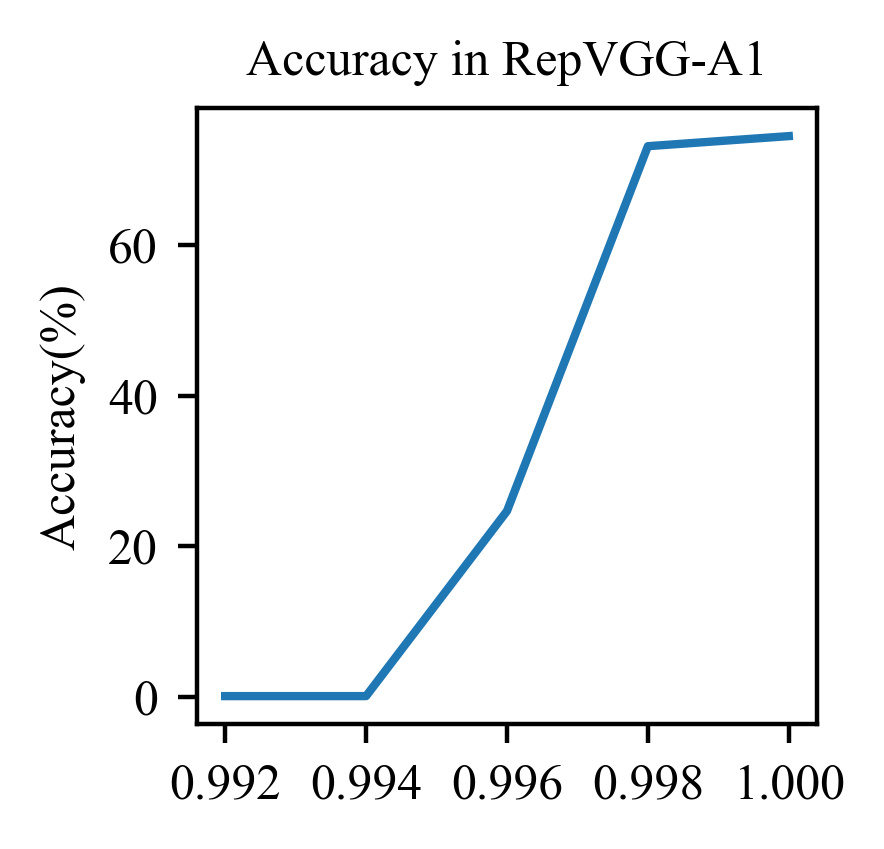}
        \end{minipage}\label{fig:repvgg_a1_w_lines20}
    }
    \subfigure[$21^{st}$]{
        \begin{minipage}{0.18\textwidth}
            \centering
            \includegraphics[scale=0.6, trim=20 0 0 17, clip]{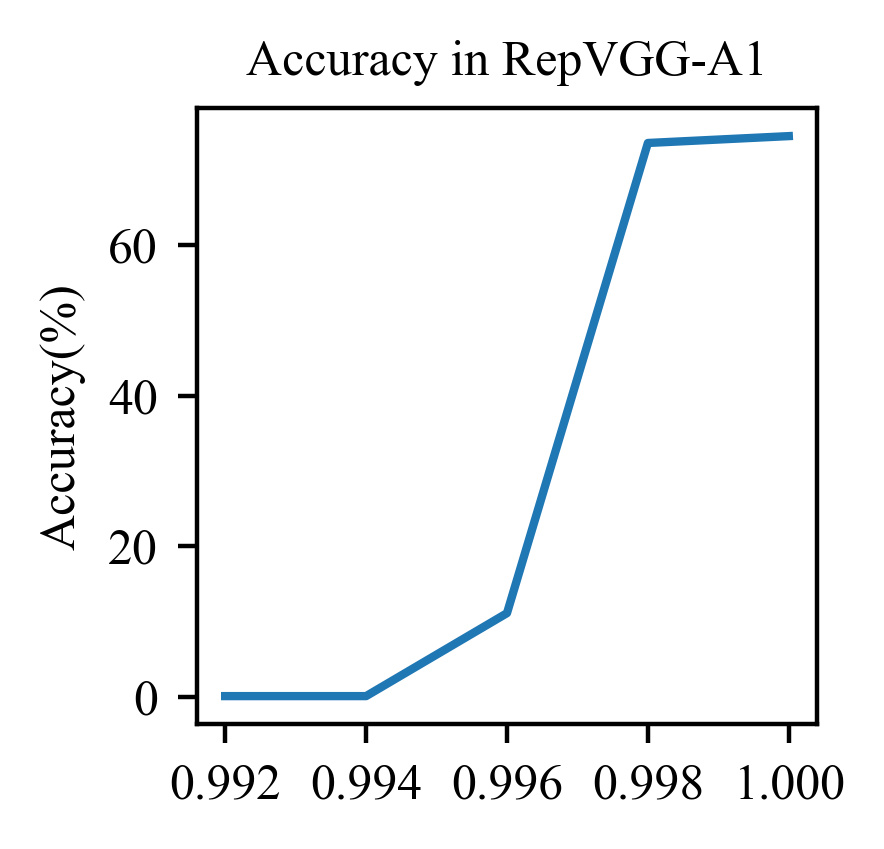}   
        \end{minipage}\label{fig:repvgg_a1_w_lines19}
    }
    \caption{The influence of weight clipping (not quantization), we clip the weights within $[-\alpha \max(|W|), \alpha \max(|W|)]$ on $19^{th}\sim21^{st}$ conv layers.}
    \label{fig:weight_line}
\end{figure}

We further investigate the problem based on the conversion process (as illustrated in Figure \ref{fig:conversion_kernels}) where the weights from the identity branch and the 1×1 Conv only influence the central points within the merged 3×3 kernel. Specifically, the center points of the converted kernel exhibit a wide range of values, while the surrounding weights (i.e., the eight non-central points in a 3×3 matrix) have a smaller range (for more details, refer to Figure \ref{fig:boxplot}). We have discovered that the wide distribution range of converted conv weights primarily stems from the distribution around the center point. \cite{ding2022re} also indicates that the central points of the kernel have a higher standard deviation compared to the surrounding points. Then we analyze the sensitivity of the weights to clipping. As Figure~\ref{fig:weight_line} shows, a slight clipping of converted weights will lead to a large drop in prediction accuracy. The analysis above serves as motivation for us to seek a better quantization technique for the central weights.


\begin{figure}
    \centering
    \includegraphics[width=1\linewidth]{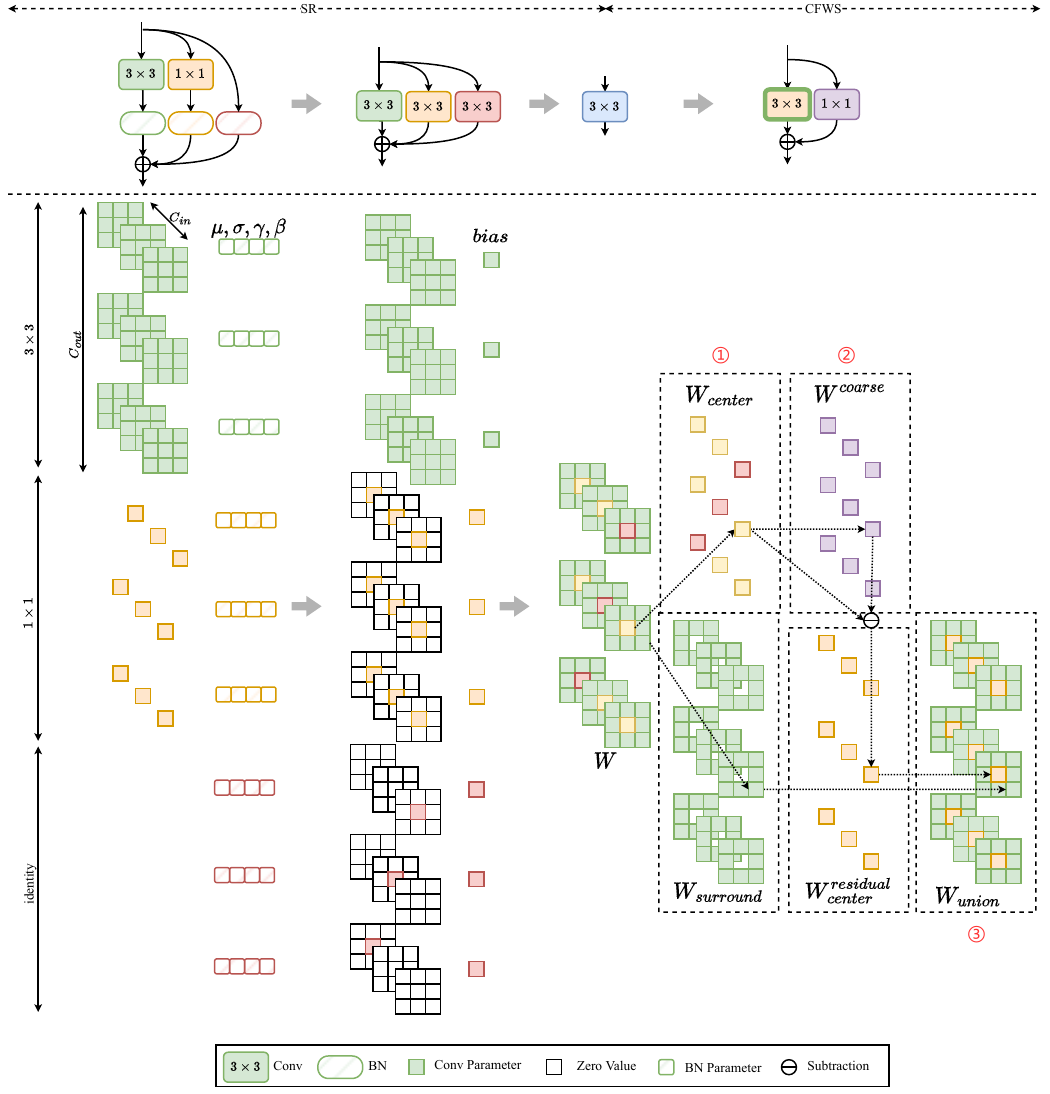}
    \caption{Diagram of the CFWS. Firstly, the multi-branch weights are fused with the batch-norm parameters. Secondly, the multi-branch structure is converted into a single-branch. Thirdly, we apply our CFWS on the single-branch model as the 3 steps: 1) split the weights $W$ into the center weight $W_{center}$ and the surrounding areas $W_{surround}$. 2) Perform a coarse quantization of $W_\text{center}$. 3) Perform a fine quantization of $W_\text{union}$.}
    \label{fig:cfws}
\end{figure}

To solve the problem, we propose Coarse \& Fine Weight Splitting (CFWS) to quantize the center weights and the surrounding weights separately.
Given the weights of the converted conv, $W \in \mathbb{R}^{C_{out} \times C_{in} \times 3 \times 3}$, where $C_{out}$ and $C_{in}$ represent the number of output and input channels, respectively,  The overall process can be viewed as:
\begin{equation}
W \approx \mathcal{W} = \mathcal{W}^{coarse} + \mathcal{W}^{fine}
\end{equation}
As shown in Figure \ref{fig:cfws}, our CFWS mainly consists of 3 steps:

\textbf{1. Split the weights.}
First, we split the weights into two categories: (a) center weights $W_{\text{center}} \in \mathbb{R}^{C_{out} \times C_{in} \times 1 \times 1}$; (b) the non-central weights $W_{\text{surround}} \in \mathbb{R}^{C_{out} \times C_{in} \times 3 \times 3}$ where the center points are empty.

\textbf{2. Perform a coarse quantization of $W_\text{center}$.} 
Second, we perform a coarse quantization of $W_\text{center}$:
\begin{equation}
\mathcal{W}_{\text{center}} = Q(W_{\text{center}}, s^{\text{coarse}})\cdot s^{\text{coarse}}
\end{equation}
where the scale factor $s^{\text{coarse}}$ is computed by standard Min-Max~\cite{jacob2018quantization} calibration:
\begin{equation}
    s^{\text{coarse}} = \frac{\max(|W_{\text{center}}|)}{2^{k-1}-1}.
\end{equation}
Since this step involves a coarse quantization $Q(W_{\text{center}}^{\text{coarse}}, s^{\text{coarse}})$, it is necessary to compensate for the residual which is:
\begin{equation}
W_{\text{center}}^{\text{residual}} = W_{\text{center}} - \mathcal{W}_{\text{center}}
\end{equation}

\textbf{3. Perform a fine quantization of $W_\text{union}$.}
From a global perspective, the second step can be viewed as a coarse quantization of $W$. In other words, we now have $\mathcal{W}^{coarse} = \mathcal{W}_{\text{center}}$ and all the remaining weights $W_\text{union} = \text{Union}(W_{\text{center}}^{\text{residual}}, W_\text{surround})$. Therefore, the last step is to apply a fine quantization on $W_\text{union}$:
\begin{equation}
\mathcal{W}^{fine} = Q(W_\text{union}, s^{\text{fine}})\cdot s^{\text{fine}} 
\end{equation}
where the scale factor $s^{\text{fine}}$ is also computed by:
\begin{equation}
s^{\text{fine}}=\frac{\max(|W_{\text{mix}}|)}{2^{k-1}-1}.
\end{equation}

\begin{figure}
    \centering
    \subfigure[$18^{th}$ $W^{\text{coarse}}$]{
        \begin{minipage}{0.2\textwidth}
            \centering
            \includegraphics[scale=0.65,  trim=8 4 0 18, clip]{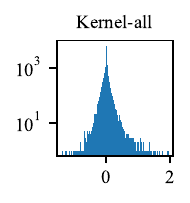}
        \end{minipage}\label{fig:18center_hista}
    }
    \subfigure[$20^{th}$ $W^{\text{coarse}}$]{
        \begin{minipage}{0.2\textwidth}
            \centering
            \includegraphics[scale=0.65,  trim=8 4 0 18, clip]{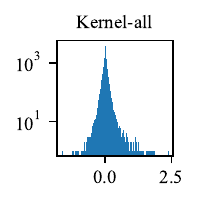}   
        \end{minipage}\label{fig:20center_hista}
    }
    \subfigure[$22^{nd}$ $W^{\text{coarse}}$]{
        \begin{minipage}{0.2\textwidth}
            \centering
            \includegraphics[scale=0.65,  trim=8 4 0 18, clip]{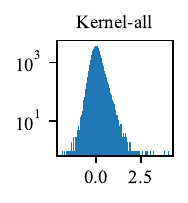}
        \end{minipage}\label{fig:22center_hista}
    }
    \\
    \subfigure[$18^{th}$ $W_{\text{center}}^{\text{residual}}$]{
        \begin{minipage}{0.2\textwidth}
            \centering
            \includegraphics[scale=0.65,  trim=8 4 0 18, clip]{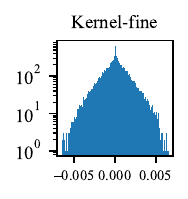}
            
        \end{minipage}\label{fig:18center_histb}
    }
    \subfigure[$20^{th}$ $W_{\text{center}}^{\text{residual}}$]{
        \begin{minipage}{0.2\textwidth}
            \centering
            \includegraphics[scale=0.65,  trim=8 4 0 18, clip]{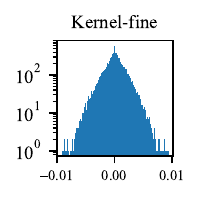}
            
        \end{minipage}\label{fig:20center_histb}
    }
    \subfigure[$22^{nd}$ $W_{\text{center}}^{\text{residual}}$]{
        \begin{minipage}{0.2\textwidth}
            \centering
            \includegraphics[scale=0.65,  trim=8 4 0 18, clip]{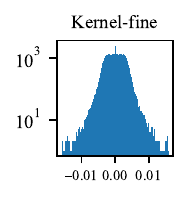}
            
        \end{minipage}\label{fig:22center_histb}
    }\\
    \subfigure[$18^{th}$ $W_{\text{surround}}$]{
        \begin{minipage}{0.2\textwidth}
            \centering
            \includegraphics[scale=0.65,  trim=8 4 0 18, clip]{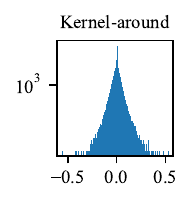}
        \end{minipage}\label{fig:18center_histc}
    }
    \subfigure[$20^{th}$ $W_{\text{surround}}$]{
        \begin{minipage}{0.2\textwidth}
            \centering
            \includegraphics[scale=0.65,  trim=8 4 0 18, clip]{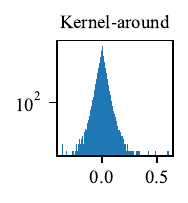}
        \end{minipage}\label{fig:20center_histc}
    }
    \subfigure[$22^{nd}$ $W_{\text{surround}}$]{
        \begin{minipage}{0.2\textwidth}
            \centering
            \includegraphics[scale=0.65,  trim=8 4 0 18, clip]{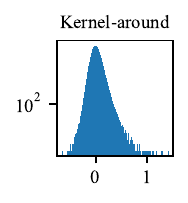}
        \end{minipage}\label{fig:22center_histc}
    }\\
    \subfigure[$18^{th}$ $W_{\text{union}}$]{
        \begin{minipage}{0.2\textwidth}
            \centering
            \includegraphics[scale=0.65,  trim=8 4 0 18, clip]{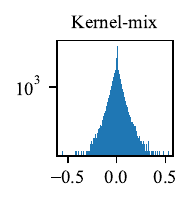}
        \end{minipage}\label{fig:18center_histd}
    }
    \subfigure[$20^{th}$ $W_{\text{union}}$]{
        \begin{minipage}{0.2\textwidth}
            \centering
            \includegraphics[scale=0.65,  trim=8 4 0 18, clip]{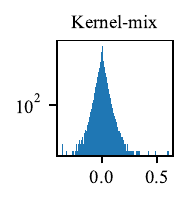}
        \end{minipage}\label{fig:20center_histd}
    }
    \subfigure[$22^{nd}$ $W_{\text{union}}$]{
        \begin{minipage}{0.2\textwidth}
            \centering
            \includegraphics[scale=0.65,  trim=8 4 0 18, clip]{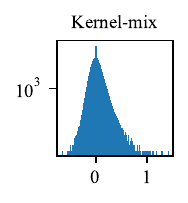}
        \end{minipage}\label{fig:22center_histd}
    }\\
    \caption{the distribution of $W^{\text{coarse}}$, $W_{\text{center}}^{\text{residual}}$, $W_{\text{surround}}$ and $W_{\text{union}}$ of the $18^{th}, 20^{th}$ and $ 22^{nd}$ layers during CFWS.} 
    \label{fig:center_hist}
\end{figure}

We show the distribution of $W^{\text{coarse}}, W^{\text{residual}}_{center}, W_{\text{surround}}, W_{\text{center}}$ in Figure \ref{fig:center_hist} and have two observations: (1) a considerable disparity in the range between $W_{\text{surround}}$ and $W_{\text{center}}$; (2) the majority of values are concentrated within a narrow range near zero, with only a few values being outliers.
By employing our proposed CFWS, the quantization errors of both $W_{\text{center}}$ and $W_{\text{surround}}$ are significantly reduced. For $W_{\text{center}}$, the coarse quantization error is compensated by the subsequent fine quantization. As for $W_{\text{surround}}$, its quantization range is adjusted to mitigate the influence of large values in $W_{\text{center}}$.

To evaluate the efficiency of our proposed CFWS without relying on a specific target platform, we adopt Bit Operations (BOPs) \cite{van2020bayesian, van2022simulated, prasad2023practical} as a surrogate metric. BOPs are defined as follows:
\begin{equation}
BOPS(\phi) = \sum_{\text op_i \in \text{network}} bits\left(\phi_{i}\right) MAC\left(op_{i}\right)
\end{equation}
where $op_i$ represents the operations performed in the network, $bits(\phi_{i})$ represents the bit-width associated with weights and activations for the operation opi, and $MAC(op_i)$ represents the total number of Multiply-Accumulate (MAC) operations for the operation $op_i$.
In Table \ref{tab:16bit}, it is worth noting that our method exhibits substantial improvements in accuracy compared to INT8 quantization, with minimal increase in bit-operations (BOPs). Furthermore, when compared to INT16 quantization, our method not only achieves higher accuracy but also requires significantly fewer BOPs.

\subsection{Quantization Metric for Activation}\label{sec:kl}
When analyzing the activation distribution, we introduce a method to analyze the influence of activation clipping.
We manually set the clipping range of values of the activations for each conv layer and observe the effect on the accuracy of the network on the validation set from ImageNet by changing the range of clipping.
This allows us to understand the importance of activations with different values.
It is implemented as follows:
\begin{enumerate}
    \item Input a number of input samples to the network, and collect the activation values $A$ of each layer. 
    \item Compute the the max absolute value $\max(|A|)$ of $A$. 
    \item Clip the $A$ into a range of $[-\alpha \max(|A|), \alpha \max(|A|)]$. Data within this range is retained, and the outside part is clipped as $\pm\alpha \max(|A|)$.
    \item Shrink the range by decreasing $\alpha$ step by step. In each step, collect the prediction accuracy of the network.
\end{enumerate}

\begin{table}
\centering
\caption{Comparison between standard 8/16-bit PTQ and Our 8-bit PTQ}
\label{tab:16bit}
\begin{tabular}{cccc}
\hline
\rule{0pt}{12pt}
    Network  & PTQ & \thead{Top-1 Acc (\%)} & \thead{BOPs (G)} \\
\hline
\rule{0pt}{12pt}
    RepVGG-A1 & TensorRT 8-bit  \cite{davoodi2019tensorrt}      &  64.0  & 151.7\\
              & our CFWS  &  \textbf{74.2}  & 168.4\\
              & TensorRT 16-bit \cite{davoodi2019tensorrt}     &  72.8  & 606.1\\
\hline
\rule{0pt}{12pt}
    RepVGG-B1 & TensorRT 8-bit \cite{davoodi2019tensorrt}     &  26.8  & 756.5\\
              & our CFWS   &  \textbf{75.7}  & 840.8\\
              & TensorRT 16-bit  \cite{davoodi2019tensorrt}     &  72.4  & 3027.0\\
\hline
\rule{0pt}{12pt}
\end{tabular}
\end{table}

\begin{figure}
\centering
    
\subfigure[$19^{th}$]{
    \begin{minipage}{0.3\textwidth}
        \centering
        \includegraphics[scale=0.5, trim=0 0 0 17, clip]
        {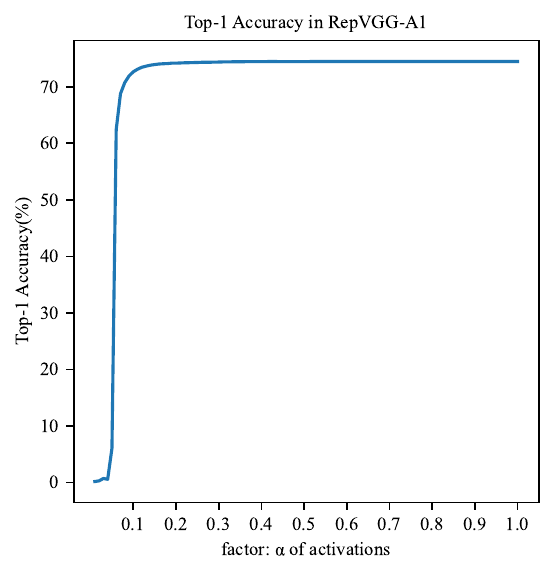}
    \end{minipage}\label{fig:repvgg_a1_lines19}
}\hfill
\subfigure[$20^{th}$]{
    \begin{minipage}{0.3\textwidth}
        \centering
        \includegraphics[scale=0.5, trim=0 0 0 17, clip]
        {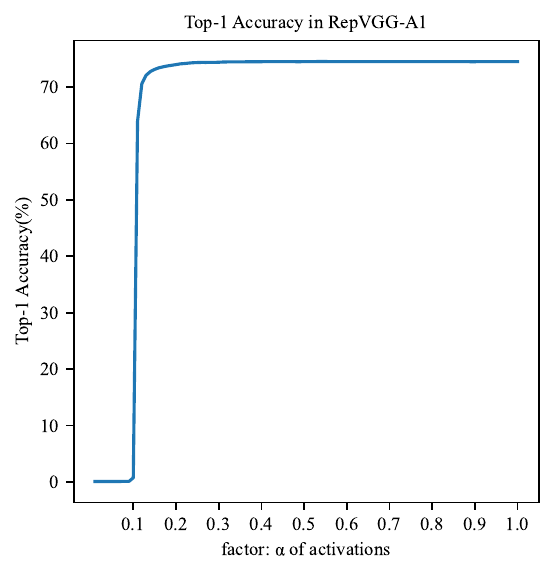}
    \end{minipage}\label{fig:repvgg_a1_lines20}
}\hfill
\subfigure[$21^{st}$]{
    \begin{minipage}{0.3\textwidth}
        \centering
        \includegraphics[scale=0.5, trim=0 0 0 17, clip]
        {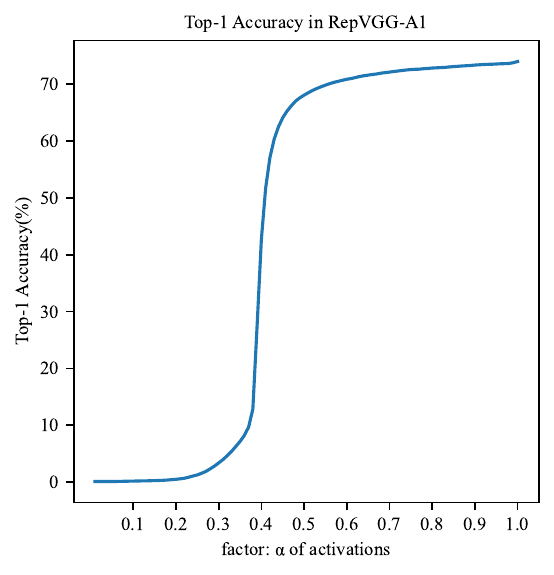}   
    \end{minipage}\label{fig:repvgg_a1_lines21}
}
\caption{The influence of activation clipping (not quantization), we clip the activation within $[-\alpha \max(|A|), \alpha \max(|A|)]$ on $19^{th}\sim21^{st}$ conv layers.}
\label{fig:repvgg_a1_lines}
\end{figure}

The $\alpha$-accuracy curve of the network is plotted in Figure~\ref{fig:repvgg_a1_lines}.
We can find that the large values of the activation value distribution do not affect the overall accuracy of the network, which can be even improved after proper clipping. 
In other words, there are a large number of outliers that are insensitive to clipping in activations. 
If a proper clipping method can be adopted, we can retain the essential information and discard outliers. 
This clipping will narrow the quantization range of the quantization so as to reduce the quantization error.

According to this observation, the Min-Max quantization is not appropriate for the quantization of activation.
To reduce the quantization range, we use a metric $L$ to evaluate the distance of the tensors before and after quantization, which is formulated as:
\begin{equation}
    \arg\mathop{\min}_{s}L(A, Q(A,s) \cdot s).
\end{equation}

There are various metrics to evaluate the distance of two tensors, such as mean-square-error (MSE), cosine distance, and KL distance.
Previous PTQ methods for reparameterization-based networks use the MSE as the metric to quantize the activations.
However, we find the MSE metric is not optimal.
According to the finding of the above analysis for the distribution of the activations, there are many large values that are not sensitive to clipping.
However, clipping these large values will cause a large MSE between the activations before and after quantization.
Therefore, the quantization range using MSE metric is large, which is not optimal.
We explore different metrics and demonstrate the results in table~\ref{tab:act_metrics}.
We observe that the KL metric is better than other metrics.

\begin{table}
\centering
\caption{Comparison of different existing metrics used on activation quantization.Top-1 accuracies and the scaling factors $s$ in $18^{th}\sim 20^{th}$ conv layers of RepVGG-A1 when using Min-Max~\cite{jacob2018quantization}, MSE~\cite{nagel2021white}, Cosine~\cite{wu2020easyquant} distance and KL from TensorRT~\cite{davoodi2019tensorrt} distance for Activation Quantization are shown in the table. As for the weight, we use Min-Max quantization.}
\label{tab:act_metrics}
\begin{tabular}{ccccc}
\hline
\rule{0pt}{12pt}
Metrics & \thead{Top-1\\Acc (\%)}  & \thead{$s$ in\\$18^{th}$ layer} & \thead{$s$ in\\$19^{th}$ layer} & \thead{$s$ in\\$20^{th}$ layer}\\
\hline
\rule{0pt}{12pt}
Min-Max & 2.7 & 1.601 & 0.932 & 0.710\\
MSE     & 6.6 & 0.529 & 0.539 & 0.464\\
Cosine  & 2.6 & 0.106 & 0.122 & 0.100\\
KL      & \textbf{64.0} & 0.208 & 0.127 & 0.098\\
\hline
\rule{0pt}{12pt}
\end{tabular}
\end{table}

The original KL metric is proposed in TensorRT.
However, we observed that there is a numerical problem in computing the KL-divergence, which is formulated as follows:
\begin{equation}
\resizebox{.91\linewidth}{!}{$
\displaystyle
D_{\text{KL}}(dist_{fp} \parallel dist_{q})=\sum _{x\in {\mathcal {X}}}dist_{fp}(x)\log \left({\frac {dist_{fp}(x)}{dist_{q}(x)}}\right)
$}
\end{equation}
where $dist_{fp}$ is the distribution of full-precision activation, $dist_{q}$ is the distribution of quantized activation, and $\mathcal {X}$ is the quantization range.
Because the distribution values are represented as floating-point numbers, there is a numerical problem in computing $\log \left({\frac {dist_{fp}(x)}{dist_{q}(x)}}\right)$.
Sometimes the result will have a significant error, making the calibration of quantization not accurate.
To mitigate this problem, we propose to transform the computation:
\begin{equation}
    \log \left({\frac {dist_{fp}(x)}{dist_{q}(x)}}\right)=\log (dist_{fp}(x)) -\log(dist_{q}(x)).
\end{equation}
As shown in Table~\ref{tab:kl_exps}, the transformed computation achieves better quantization results.

\begin{table}
\centering
\caption{Ablation studies for activations with RepVGG A1 and B1 with KL quantization metric. The KL quantization without transformed formula or ReLU fusion is based on the idea of TensorRT~\cite{davoodi2019tensorrt}.}
\label{tab:kl_exps}
\begin{tabular}{ccccc}
\hline
\rule{0pt}{12pt}
Network    &  \thead{Transformed\\Formula}   & \thead{ReLU\\Fusion}   & \thead{Top-1\\Acc (\%)}\\
\hline
\rule{0pt}{12pt}
\thead{RepVGG-A1}  &                  &                  & 64.0\\ 
&  $\surd$         &                  & 64.8\\
&  $\surd$         & $\surd$          & \textbf{65.8}\\
\hline
\rule{0pt}{12pt}
\thead{RepVGG-B1}  
&                  &                  & 26.8\\ 
&  $\surd$         &                  & 44.5\\
&  $\surd$         & $\surd$          & \textbf{62.0}       \\
\hline
\rule{0pt}{12pt}
\end{tabular}
\end{table}

It is common to fuse the ReLU after the conv layer, meaning that the negative part of the activations from conv is clipped. 
We find that it is better to use the distribution after ReLU to compute the KL metric.
Table \ref{tab:kl_exps} shows the results of using transform and relu fusion for KL-based quantization.

\section{Experimental Evaluation}
To evaluate the proposed methods, we experiment with different reparameterization-based networks.
We will introduce the experimental settings.
Then we will demonstrate the results on networks for both classification and object detection, respectively.
At last, we will show our ablation studies.

\subsection{Settings}

{We conducted our experiments on NVIDIA RTX3090.
For calibration during quantization, only 32 unlabeled calibration samples were used.}

We experiment on RepVGG networks~\cite{ding2021repvgg} with different sizes, including RepVGG-A0, RepVGG-A1, and RepVGG-B1, for ImageNet~\cite{deng2009imagenet} classification task.
The input of RepVGG networks is RGB images with a resolution of 224x224.
We also experiment on the reparameterization-based networks for object detection task, including YOLOv6t and YOLOv6s~\cite{li2022yolov6}.
The input of YOLOv6 networks is RGB images with a resolution of 640x640.
We use 32 unlabeled calibration images to quantize the network.

We compare the proposed method with other quantization methods for the reparameterization-based model, including vanilla PTQ with KL metric~\cite{davoodi2019tensorrt}, RepOpt~\cite{ding2022re} and QARepVGG~\cite{chu2022make}.
The results of these works are taken from the original papers.

\subsection{ImageNet Classification}

We demonstrate the top-1 accuracy on ImageNet test dataset of both full-precision network (FP) and quantized networks (Quant) in Table~\ref{tab:ImageNet}.
As we use the original RepVGG network without training, the FP result is the same as the original result.
While RepOpt-VGG and QARepVGG re-train the network, resulting in a lower FP result.
For quantized networks, the results of the vanilla PTQ method significantly drop, which is not acceptable for deployment.
Benefit from re-training, RepOpt-VGG and QARepVGG achieve much higher results after quantization.
{The degrees of decline in accuracy after quantization can be used for a fair comparison.
This comparison shows that }our method also achieves a significant improvement for quantized networks, which is comparable with the training-based quantization methods.
For example, we achieve 69.2\% top-1 accuracy on RepVGG-A0, which is much better than vanilla PTQ (52.2\%), slightly better than RepOpt-VGG (64.8\%), and slightly worse than QARepVGG (70.4\%).

\begin{table}
\centering
\caption{The top-1 accuracy (\%) on ImageNet when applying quantization on RepVGG. The PTQ method refers to applying the standard PTQ approach to the model, utilizing the KL metric from TensorRT.\cite{davoodi2019tensorrt}}
\label{tab:ImageNet}
\begin{threeparttable}
\begin{tabular}{ccccc}
    \hline
\rule{0pt}{12pt}
    Network & method & \thead{retraining} & FP & Quant \\
    \hline
\rule{0pt}{12pt}
    RepVGG-A0 & PTQ \cite{davoodi2019tensorrt}  & - & 72.4  & 52.2(20.2$\downarrow$)\\
    &RepOpt$^{*} $  \cite{ding2022re} & $\surd$ & 70.3  & 64.8(5.5$\downarrow$)\\
    &QARepVGG$^{*} $  \cite{chu2022make} & $\surd$ & 72.2  & 70.4(1.8$\downarrow$)\\
    &Ours  & - & 72.4  & 69.2(3.2$\downarrow$)\\
    \hline
\rule{0pt}{12pt}
RepVGG-A1 & PTQ \cite{davoodi2019tensorrt}  & - & 74.5  & 61.7(12.8$\downarrow$)\\
    &RepOpt$^{*} $  \cite{ding2022re} & $\surd$ & -     & -\\
    &QARepVGG$^{*} $  \cite{chu2022make} &  $\surd$  & -     & -\\
    &Ours  &  -  & 74.5 & 74.2(0.3$\downarrow$)\\
    \hline
\rule{0pt}{12pt}
    RepVGG-B1&PTQ  \cite{davoodi2019tensorrt} &  -  & 78.4  & 54.6(23.8$\downarrow$)\\
    &RepOpt$^{*}$  \cite{ding2022re} &  $\surd$  & 78.5  & 75.9(2.6$\downarrow$)\\
    &QARepVGG$^{*}$  \cite{chu2022make} &  $\surd$  & 78.0  & 76.4(1.6$\downarrow$)\\
    &Ours  &  -  & 78.4  & 75.7(2.7$\downarrow$)\\
    \hline
\rule{0pt}{12pt}
\end{tabular}
\vspace{-1em}  
\begin{tablenotes}
\scriptsize{\item $*$: These methods necessitate modifications to the network architectures and retraining.}
\end{tablenotes}
\end{threeparttable}
\end{table}

\subsection{COCO Object Detection}

Similarly, on COCO object detection test dataset, most of the mAPs \cite{li2019fully} for the YOLOv6, its RepOptimizer method, and its QARepVGG method are obtained from \cite{chu2022make}. 
Besides, the performances of YOLOv6s and its RepOptimizer method are provided by the Meituan official. 
And the experiments in Table \ref{tab:coco} demonstrate that our method has generalization on large target detection reparameterization-based networks.
For instance, we also achieve 43.0\% mAP on YOLOv6s with a drop of 0.8\%, which far exceeds the vanilla PTQ (41.2\%) with a drop of 2.6\%. 
We also noted that QARepVGG is not optimal for all of the networks for a drop of 1.3\%.

\begin{table}
\centering
\caption{The mAP (\%) on COCO of existing quantization methods. The PTQ method refers to applying the standard PTQ approach to the model, utilizing the KL metric from TensorRT.\cite{davoodi2019tensorrt}}
\label{tab:coco}
\begin{threeparttable}
\begin{tabular}{ccccc}
\hline
\rule{0pt}{12pt}
Network & method & \thead{retraining} & FP & Quant \\
\hline
\rule{0pt}{12pt}
YOLOv6t&PTQ \cite{davoodi2019tensorrt}    &  -  & 41.1                      & 37.8 (3.3$\downarrow$) \\
&RepOpt$^{*}$ \cite{ding2022re} &  $\surd$  & 40.7                      & 39.1 (1.6$\downarrow$) \\
&QARepVGG$^{*}$ \cite{chu2022make} & $\surd$ & 40.7                      & 39.5 (1.2$\downarrow$) \\
&Ours  &  -  & 41.1                      & 39.6 (1.5$\downarrow$) \\
\hline
\rule{0pt}{12pt}
YOLOv6s&PTQ \cite{davoodi2019tensorrt} &  -  & 43.8                      & 41.2 (2.6$\downarrow$) \\
&RepOpt$^{*}$ \cite{ding2022re} &  $\surd$  & 43.1                      & 42.6 (0.5$\downarrow$) \\
&QARepVGG$^{*}$  \cite{chu2022make} &  $\surd$  & 42.3                      & 41.0 (1.3$\downarrow$) \\
&Ours &  -  & 43.8                      & 43.0 (0.8$\downarrow$) \\
\hline
\rule{0pt}{12pt}
\end{tabular}
\vspace{-1em}  
\begin{tablenotes}
\scriptsize{\item $*$: These methods necessitate modifications to the network architectures and retraining.}
\end{tablenotes}
\end{threeparttable}
\end{table}

\subsection{Ablation Studies}

We conduct ablation studies on the proposed methods, which are shown in Table \ref{tab:kl_cfws}.
We observe that both transformed KL quantization metric with ReLU fusion and CFWS can achieve significant improvement.
For example, the top-1 accuracy is lifted to 62.0\% after using the KL metric to clip the distribution of activations on RepVGG-B1. 
After conducting the CFWS on the distribution of weights, the prediction accuracy is improved to 54.2\%. 
And then, we further improve the performance (75.7\%) of the quantized reparameterization-based networks with the two proposed methods together.

\begin{table}
\centering
\caption{Ablation studies for proposed PTQ method with RepVGG A1 and B1.}
\label{tab:kl_cfws}
\begin{tabular}{ccccc}
\hline
\rule{0pt}{12pt}
Network    &  \thead{KL metric \\with ReLU fusion}   & \thead{CFWS}   & \thead{Top-1\\Acc (\%)}\\
\hline
\rule{0pt}{12pt}
\thead{RepVGG-A1}         
&                  &                  & 64.0\\ 
&  $\surd$         &                  & 65.8\\
&                  & $\surd$          & 73.1\\
&  $\surd$         & $\surd$          & \textbf{74.2}\\
\hline
\rule{0pt}{12pt}
\thead{RepVGG-B1}
&                  &                  & 26.8\\ 
&  $\surd$         &                  & 62.0\\
&                  & $\surd$          & 54.2\\
&  $\surd$         & $\surd$          & \textbf{75.7}       \\
\hline
\rule{0pt}{12pt}
\end{tabular}
\end{table}

\section{Conclusion}
This paper presents a novel approach for post-training quantization (PTQ) of reparameterized networks. We propose the Coarse \& Fine Weight Splitting (CFWS) method to reduce quantization error in weight distribution and develop an improved KL metric for optimal quantization scales of activations. Our method bridges the gap between reparameterization techniques and practical PTQ deployment, achieving competitive quantization accuracy with minimal accuracy loss. The CFWS method effectively addresses the challenges associated with quantizing reparameterized networks, making them more efficient and scalable for various computer vision tasks, including image classification and object detection.



\bibliography{elsarticle-template-num}





\end{document}